\documentclass[conference]{IEEEtran}
\IEEEoverridecommandlockouts
\usepackage{cite}
\usepackage{amsmath,amssymb,amsfonts}
\usepackage{algorithmic}
\usepackage{algorithm}
\usepackage{graphicx}
\usepackage{textcomp}
\usepackage{xcolor}
\usepackage{url}
\usepackage{float}
\usepackage{placeins}
\usepackage{breqn}
\usepackage{array}
\usepackage{balance}

\def\BibTeX{{\rm B\kern-.05em{\sc i\kern-.025em b}\kern-.08em
    T\kern-.1667em\lower.7ex\hbox{E}\kern-.125emX}}

\begin{document}

\title{Deep Attention-based Sequential Ensemble Learning for BLE-Based Indoor Localization in Care Facilities}

\author{\IEEEauthorblockN{Minh Triet Pham}
\IEEEauthorblockA{\textit{School of Computing and}\\
\textit{Information Systems}\\
\textit{The University of Melbourne}\\
Melbourne, Australia\\
minhtrietp@student.unimelb.edu.au}
\and
\IEEEauthorblockN{Quynh Chi Dang}
\IEEEauthorblockA{\textit{School of Computing and}\\
\textit{Information Systems}\\
\textit{The University of Melbourne}\\
Melbourne, Australia\\
quynhchid@student.unimelb.edu.au}
\and
\IEEEauthorblockN{Le Nhat Tan}
\IEEEauthorblockA{\textit{Faculty of Applied Science}\\
\textit{Ho Chi Minh City University of Technology}\\
\textit{Vietnam National University - Ho Chi Minh City}\\
Ho Chi Minh City, Vietnam\\
lenhattan@hcmut.edu.vn}
}

\maketitle

\begin{abstract}
Indoor localization systems in care facilities enable optimization of staff allocation, workload management, and quality of care delivery. Traditional machine learning approaches to Bluetooth Low Energy (BLE)-based localization treat each temporal measurement as an independent observation, fundamentally limiting their performance. To address this limitation, this paper, developed as part of the ABC 2026 Activity and Location Recognition Challenge, introduces Deep Attention-based Sequential Ensemble Learning (DASEL), a novel framework that reconceptualizes indoor localization as a sequential learning problem. The framework integrates frequency-based feature engineering, bidirectional GRU networks with attention mechanisms, multi-directional sliding windows, and confidence-weighted temporal smoothing to capture human movement trajectories. Evaluated on real-world data from a care facility using 4-fold temporal cross-validation, DASEL achieves a macro F1 score of 0.4438, representing a 53.1\% improvement over the best traditional baseline (0.2898). This breakthrough demonstrates that modeling temporal dependencies in movement patterns is essential for accurate indoor localization in complex real-world environments.
\end{abstract}


\section{Introduction}
Indoor localization systems in care facilities enable optimization of staff allocation, workload management, and quality of care delivery \cite{b1, b2}. Accurate tracking of caregiver movements provides insights into care routines, enhances hand hygiene monitoring, and supports health interventions for elderly residents with conditions such as Alzheimer's and dementia \cite{b3, b4}. Automated location recording eliminates manual logging burdens and provides objective data for facility management and quality improvement initiatives \cite{b5}.

Bluetooth Low Energy (BLE) technology has emerged as a prominent indoor localization solution due to its low cost, minimal power consumption, and ease of deployment \cite{b6}. Indoor positioning methods have evolved from simple RSSI measurements to more advanced approaches such as CSI, RTT, and AoA, increasingly combined with Machine Learning techniques \cite{b7}. Traditional approaches predominantly rely on RSSI fingerprinting with classical classification algorithms including K-Nearest Neighbors, Support Vector Machines, and Random Forest classifiers \cite{b8, b9}. However, these approaches share a fundamental limitation: they treat each temporal measurement as an independent observation, extracting statistical features from beacon signals and classifying each moment in isolation. The growing complexity of indoor environments requires solutions that can handle sensor noise, multipath fading effects, and temporal dependencies that traditional independent-window classification methods cannot adequately address \cite{b10}.

BLE-based localization in real-world care facilities confronts significant data quality challenges. BLE signals experience substantial fluctuations in RSSI values caused by multipath propagation and environmental interference \cite{b11, b12}. Care facilities present particularly complex deployment scenarios characterized by beacon placement constraints and architectural complexity \cite{b8}. These environments exhibit several documented challenges: signal instability arising from multipath effects and device heterogeneity \cite{b13}, spatial sparsity with limited beacon coverage in certain areas \cite{b11}, temporal irregularity with variable detection rates \cite{b14}, and severe class imbalance where common areas receive significantly more visits than individual patient rooms \cite{b15}. Traditional machine learning methods applied to such real-world datasets demonstrate limited performance despite various optimization efforts \cite{b9, b10}. This performance plateau reveals a critical limitation: the independence assumption inherent in traditional approaches discards temporal dependencies in human movement patterns \cite{b16}. Caregivers follow continuous trajectories through physical space, where location at time $t$ strongly predicts location at time $t+1$, and transitions between rooms produce gradual shifts in beacon signal patterns \cite{b17}. Traditional methods treat each temporal window as an isolated classification problem, thereby discarding contextual information about movement trajectories, dwell times, and transition dynamics.

This work addresses the location recognition component of the ABC 2026 Activity and Location Recognition Challenge, which evaluates methods using macro F1 score as the primary metric for location classification. Our research objectives are threefold: (\textit{1}) \textit{Methodological goal}: achieve breakthrough performance improvement by leveraging temporal dependencies in human movement trajectories that traditional independent-window classification methods fundamentally discard; (\textit{2}) \textit{Technical improvements}: address RSSI instability through robust feature representations, handle the inference challenge where sequence boundaries are unknown during real-time prediction, and effectively manage severe class imbalance across 12-18 room locations; (\textit{3}) \textit{Evaluation targets}: maximize location classification macro F1 score while maintaining balanced performance across all room classes regardless of visit frequency and achieving substantial performance improvements suitable for practical deployment in real-world care facilities.

The key contributions of this work demonstrate measurable improvements addressing these objectives: (1) a novel sequential framework applying deep bidirectional recurrent networks with attention mechanisms to BLE-based care facility localization, explicitly modeling temporal movement trajectories rather than treating measurements as independent observations, (2) frequency-based representation addressing RSSI instability challenges inherent in real-world deployments, providing device-agnostic features robust to signal fluctuations, (3) two-level hierarchical ensemble combining multi-seed variance reduction with multi-directional positional robustness to handle unknown sequence boundaries during inference, (4) comprehensive baseline evaluation demonstrating that traditional paradigm optimization yields diminishing returns with performance plateauing at 0.2898 macro F1, and (5) DASEL framework achieving 0.4438 macro F1 score (53.1\% improvement), with balanced performance across all room classes and robustness to device heterogeneity, suitable for real-world deployment.

\section{Methodology}
\FloatBarrier

We propose and compare two distinct approaches (Section~\ref{subsec:tradml} and Section~\ref{subsec:dasel}) representing fundamentally different paradigms in handling temporal sensor data. The traditional machine learning approach treats each temporal window independently, extracting statistical features from RSSI values and applying gradient boosting classification. In contrast, the Deep Attention-based Sequential Ensemble Learning (DASEL) framework reconceptualizes indoor localization as a sequential learning problem, leveraging the inherent temporal continuity of human movement patterns through deep learning architectures combining frequency-based feature engineering, bidirectional recurrent networks with attention mechanisms, multi-directional sliding windows, and multi-model ensemble learning.

\subsection{Dataset}
The dataset was provided by Kyushu Institute of Technology in conjunction with the ABC 2026 Activity and Location Recognition Challenge \cite{b18, b26}. The data collection involves 25 BLE beacons strategically installed throughout the 5th floor of a care facility. Each beacon continuously transmits signals that are detected by a mobile phone carried by a caregiver (User ID 90) who moves around the floor during normal work activities. The mobile phone records RSSI values from all detectable beacons with second-level temporal precision. Concurrently, an observer (User ID 97) manually tracks and annotates the caregiver's location during specific time periods, providing ground truth labels for supervised learning.

\begin{figure}[!htbp]
\centering
\includegraphics[width=\columnwidth]{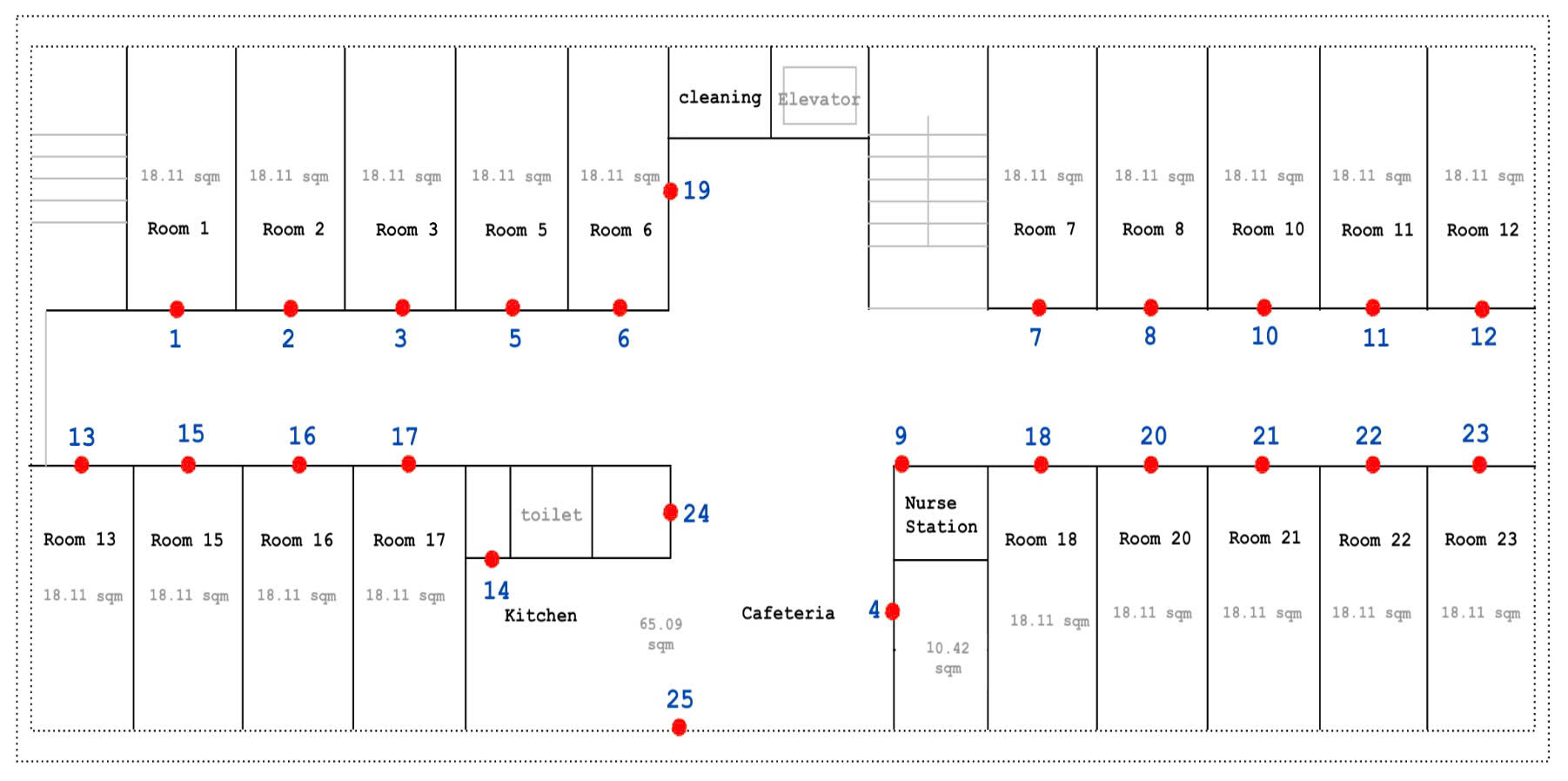}
\caption{5th floor map with beacon positions and room layout \cite{b20, b26}.}
\label{fig:floormap}
\end{figure}

\begin{figure}[!htbp]
\centering
\includegraphics[width=\columnwidth]{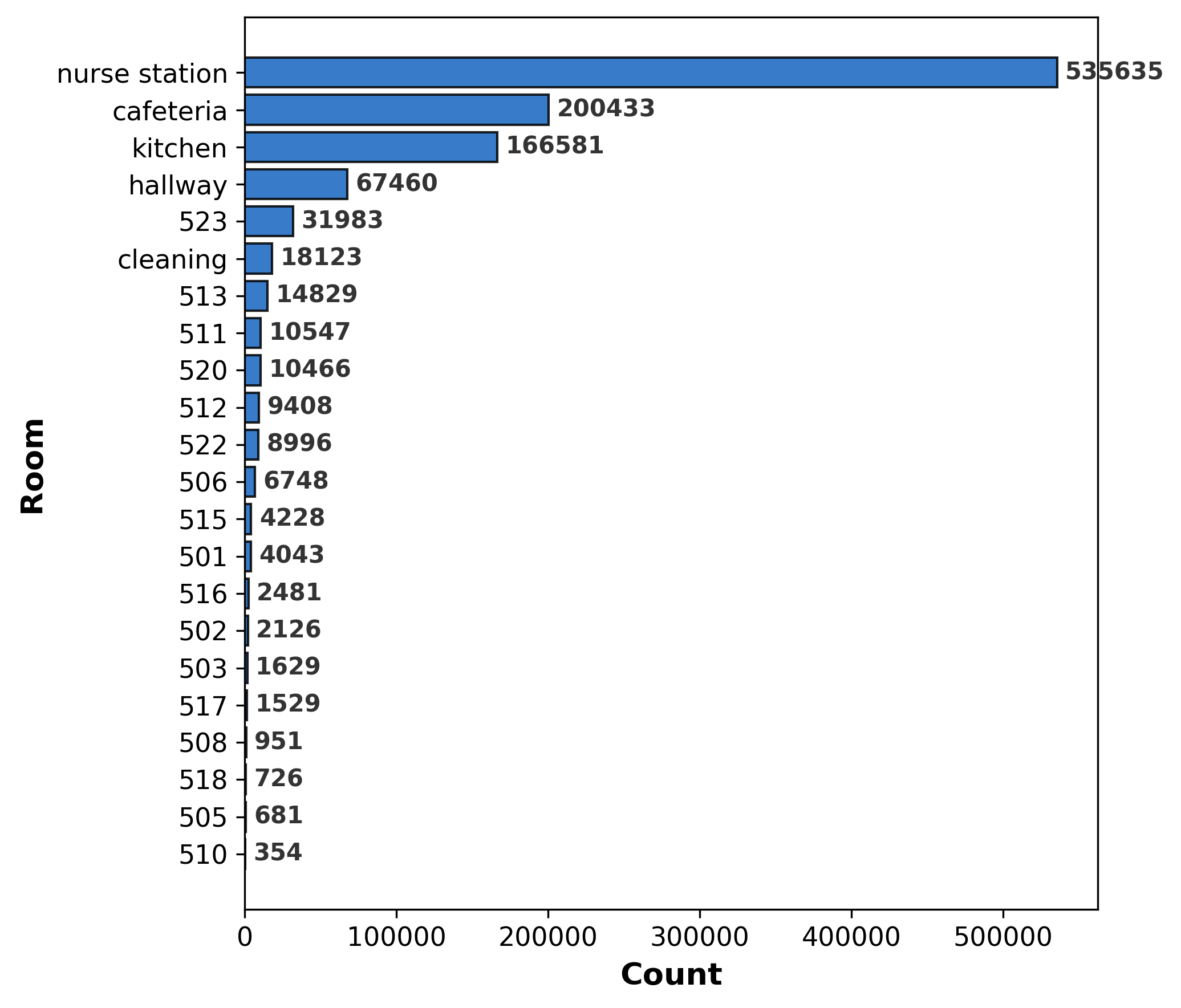}
\caption{Overall class distribution across all four days.}
\label{fig:classdist}
\end{figure}

The complete dataset spans four consecutive days (April 10-13, 2023) and covers 13-18 distinct rooms depending on the day, reflecting realistic deployment conditions where different days involve visits to different subsets of locations. After preprocessing and temporal alignment, the final labeled dataset contains approximately 1.1 million timestamped BLE sensor readings, each associated with a room label. The beacons return RSSI measurements in dBm, where values closer to zero represent stronger signals. Fig.~\ref{fig:floormap} shows the spatial layout of the 5th floor with beacon placement, while Fig.~\ref{fig:classdist} illustrates the overall class distribution across all four days.

The dataset exhibits severe class imbalance, with common areas such as nurse station and cafeteria visited far more frequently than individual patient rooms. This imbalance poses significant challenges for classification models and justifies the use of macro F1 score as the evaluation metric, ensuring balanced performance across all locations regardless of visit frequency.

\FloatBarrier
\subsection{Preprocessing}
The preprocessing pipeline transforms raw BLE sensor data and manual location annotations into a clean, labeled dataset suitable for machine learning. This step is essential to address data quality issues, ensure temporal alignment between sensor readings and labels, and create a supervised learning dataset with reliable ground truth.

The BLE sensor data from User ID 90 spanning the investigation period (April 10-13, 2023) was extracted from individual CSV files. Beacon signals were filtered to retain only transmissions from the 25 primary beacons installed on the 5th floor, excluding signals originating from beacons on other floors. MAC addresses were systematically mapped to beacon identifiers (1-25) for standardized reference. The location label data was processed to retain only valid annotations from the designated observer (User ID 97), ensuring complete temporal intervals and removing incomplete or deleted records.

Temporal alignment was performed to establish the supervised learning dataset. Each BLE sensor reading was matched with its corresponding room label based on temporal overlap between the sensor timestamp and the label's annotated time interval. All timestamps were standardized to second-level precision, consistent with established practices in indoor positioning systems where second-level temporal resolution is adequate \cite{b6, b19}. This temporal matching procedure successfully labeled approximately 1.1 million BLE records, representing 66\% of the cleaned sensor data. Records without corresponding labels, comprising 34\% of cleaned data, were excluded from the final dataset to maintain annotation quality, as these represent periods where active location tracking was not conducted or during unlabeled transitional intervals.

The final preprocessed dataset contains approximately 1.1 million labeled samples with the structure shown in Table~\ref{tab:dataset}. Each record includes a timestamp (second precision), beacon ID (1-25), RSSI value, and the corresponding room label as the prediction target.

\begin{table}[!htbp]
\caption{Final Preprocessed Dataset Structure}
\label{tab:dataset}
\centering
\renewcommand{\arraystretch}{1.5}
\begin{tabular}{|c|c|c|c|}
\hline
\textbf{timestamp} & \textbf{mac\_address} & \textbf{RSSI} & \textbf{room} \\
\hline
2023-04-10 14:21:46+09:00 & 6 & -93 & kitchen \\
\hline
2023-04-10 14:21:46+09:00 & 4 & -90 & kitchen \\
\hline
2023-04-10 14:21:46+09:00 & 4 & -97 & kitchen \\
\hline
2023-04-10 14:21:46+09:00 & 6 & -93 & kitchen \\
\hline
2023-04-10 14:21:46+09:00 & 4 & -90 & kitchen \\
\hline
... & ... & ... & ... \\
\hline
\end{tabular}
\end{table}

\FloatBarrier
\subsection{Model Training}

\subsubsection{Traditional Machine Learning Baseline}
\label{subsec:tradml}
\begin{figure}[!htbp]
\centering
\includegraphics[width=\columnwidth]{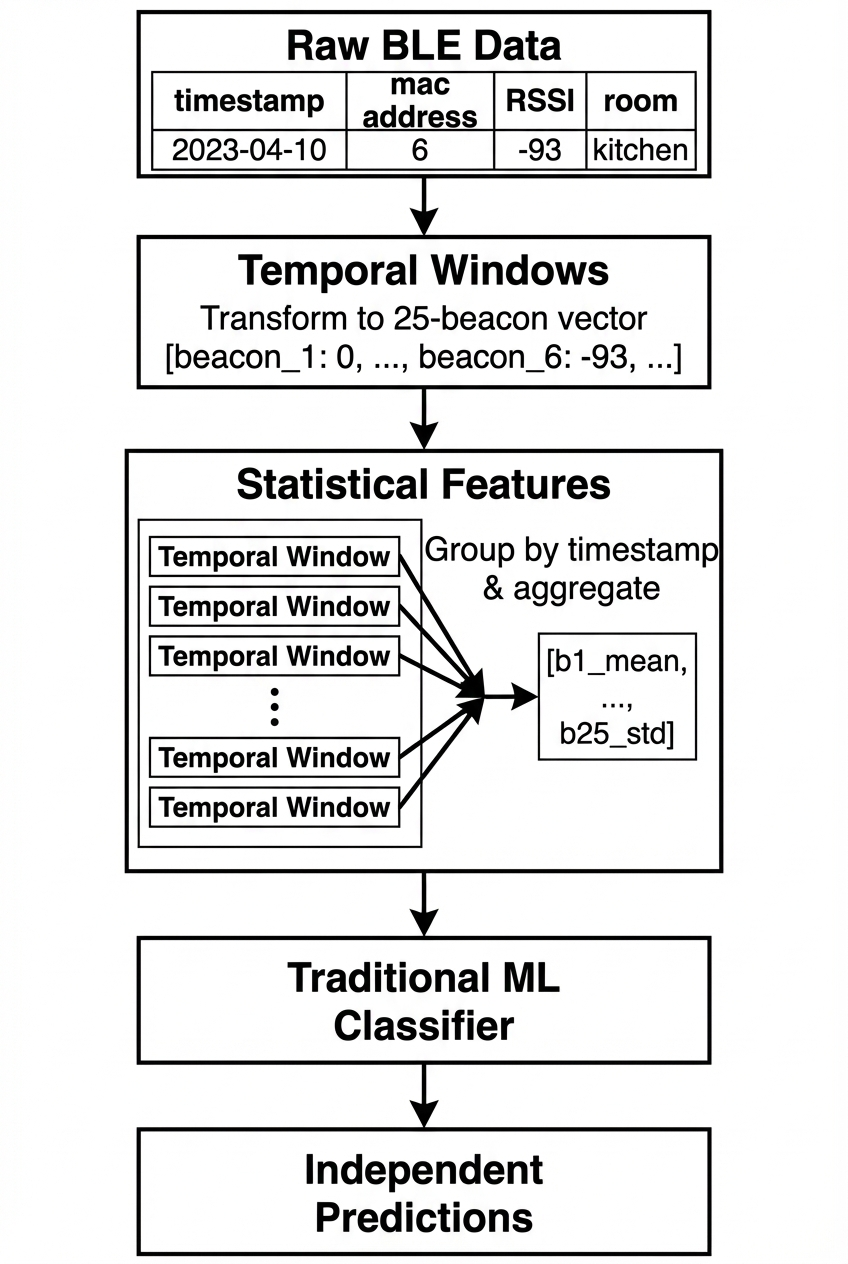}
\caption{Traditional machine learning workflow.}
\label{fig:tradml}
\end{figure}

The traditional baseline approach treats each temporal window as an independent classification problem. Fig.~\ref{fig:tradml} illustrates the complete workflow. Starting from individual BLE detection records, we construct a 25-dimensional beacon vector for each timestamp where position $i$ contains the RSSI value if beacon $i$ was detected, and 0 otherwise. Raw readings are grouped by timestamp into 1-second windows, and within each window we compute statistical features (mean, standard deviation, count) for each beacon, resulting in 75 features per window (25 beacons $\times$ 3 statistics). These aggregated features are fed into an XGBoost classifier with balanced sample weighting to address class imbalance.

To establish a comprehensive baseline, we explored optimizations from different angles. Variation 1 employed dominant beacon features by augmenting the base statistical features with three additional categorical features representing the top three most frequently detected beacons within each window, aiming to capture additional spatial context through feature engineering. Variation 2 applied signal pattern-based relabeling following Garcia and Inoue \cite{b20, b21}, using KL divergence to identify majority class rooms whose signal patterns closely match minority class rooms, then relabeling matched samples to address class imbalance through data augmentation. Despite these optimization strategies targeting both feature engineering and data handling, all traditional methods remained within a narrow performance range.

\FloatBarrier
\subsubsection{Deep Attention-based Sequential Ensemble Learning (DASEL)}
\label{subsec:dasel}
\begin{figure}[!htbp]
\centering
\includegraphics[width=\columnwidth]{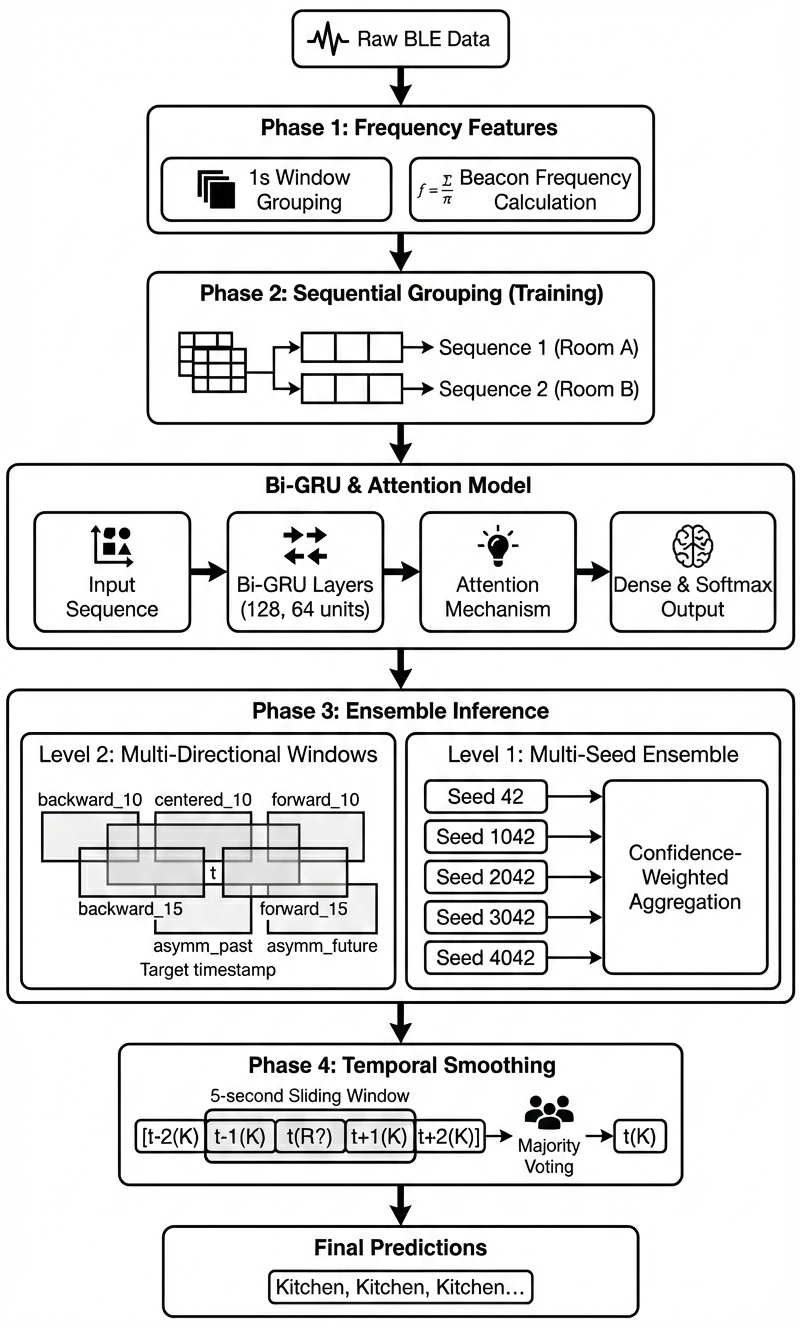}
\caption{DASEL complete workflow.}
\label{fig:daselworkflow}
\end{figure}

DASEL reconceptualizes indoor localization as a sequential learning problem through four integrated phases: (1) frequency-based feature engineering, (2) sequential model training using bidirectional recurrent networks with attention mechanisms, (3) multi-level ensemble inference combining multiple temporal perspectives and model initializations, and (4) temporal smoothing for spatial consistency. Fig.~\ref{fig:daselworkflow} provides an overview of the complete DASEL workflow.

\paragraph{Phase 1: Frequency-Based Feature Engineering}
Starting from preprocessed data (Section 2.2), we transform raw RSSI measurements into frequency-based representations. For each BLE detection record, we construct a 25-dimensional beacon vector where position $i$ contains the RSSI value if beacon $i$ was detected, and 0 otherwise. BLE readings sharing the same timestamp are grouped into 1-second windows \cite{b6, b19}. Within each window, we calculate beacon appearance frequency as:
\begin{equation}
\text{frequency}_{i,t} = \frac{\text{count}_{i,t}}{\text{total\_detections}_t}
\label{eq:frequency}
\end{equation}
where $\text{count}_{i,t}$ is the number of times beacon $i$ was detected in window $t$, and $\text{total\_detections}_t$ is the total number of all beacon detections in that window. For undetected beacons, frequency is 0. We use 23 beacons (beacons 1-23) as beacons 24-25 were never detected. Each window is represented by a 23-dimensional frequency vector with values normalized between 0 and 1.

\paragraph{Phase 2: Sequential Model Training}
During training, we segment data into sequences using ground truth room labels. We identify consecutive timestamps where the room label remains constant:
\begin{multline}
\text{room\_group\_id} = \\
\text{cumulative\_sum}(\text{room\_label} \neq \text{previous\_room\_label})
\label{eq:grouping}
\end{multline}
Each contiguous block of the same room becomes one training sequence. Sequences shorter than 3 timesteps are discarded; sequences longer than 50 timesteps are truncated by taking the last 50 timesteps.

\begin{figure}[!htbp]
\centering
\includegraphics[width=\columnwidth]{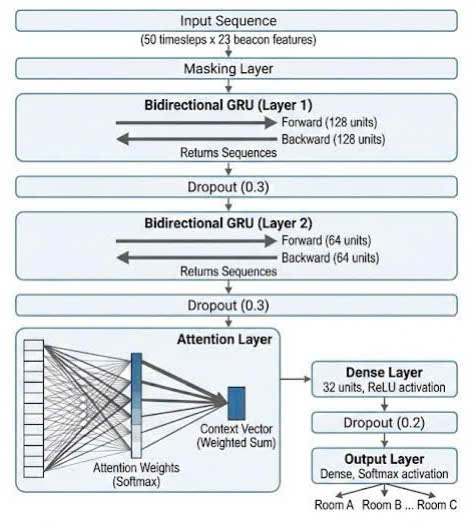}
\caption{DASEL model architecture.}
\label{fig:architecture}
\end{figure}

The model architecture (Fig.~\ref{fig:architecture}) consists of: masking layer (handles variable-length sequences, padding to 50 timesteps); first Bidirectional GRU layer (128 units); dropout (0.3); second Bidirectional GRU layer (64 units); dropout (0.3); attention mechanism with:
\begin{align}
\text{attention\_scores} &= \tanh(W \cdot \text{sequence} + b) \label{eq:attn1} \\
\text{attention\_weights} &= \text{softmax}(\text{attention\_scores}) \label{eq:attn2} \\
\text{context\_vector} &= \sum(\text{sequence} \times \text{attention\_weights}) \label{eq:attn3}
\end{align}
dense layer (32 units, ReLU, 0.2 dropout); and output layer (softmax activation). The model is trained using sparse categorical cross-entropy loss with balanced class weights.

\paragraph{Phase 3: Multi-Level Ensemble Inference}
Phase 3 implements a two-level ensemble strategy combining multi-seed model training with multi-directional sliding windows.

\textit{Level 1: Multi-Seed Model Ensemble.} We train 5 independent models with different random initialization seeds [42, 1042, 2042, 3042, 4042]. For each of the 7 directional windows, we apply probability averaging across the 5 models (Algorithm~\ref{alg:multiseed}), producing 7 probability distributions.

\begin{algorithm}[!htbp]
\caption{Multi-Seed Model Ensemble}
\label{alg:multiseed}
\begin{algorithmic}[1]
\FOR{each direction $d \in \{\text{backward\_10},$ $\text{centered\_10}, \text{forward\_10},$ $\text{backward\_15}, \text{forward\_15},$ $\text{asymm\_past}, \text{asymm\_future}\}$}
    \STATE Extract sequences from direction $d$
    \FOR{each model $i$ ($i = 1$ to $5$)}
        \STATE $\text{probability}_i = \text{model}_i.\text{predict}(\text{sequences}_d)$
    \ENDFOR
    \STATE $\text{averaged\_probability}_d = \text{mean}(\text{prob}_1, \ldots, \text{prob}_5)$
\ENDFOR
\end{algorithmic}
\end{algorithm}

\textit{Level 2: Multi-Directional Window Combination.} For each timestamp $t$ requiring prediction, we create 7 temporal windows:
\begin{itemize}
\item backward\_10: $[t-9, \ldots, t]$
\item centered\_10: $[t-4, \ldots, t+5]$
\item forward\_10: $[t, \ldots, t+9]$
\item backward\_15: $[t-14, \ldots, t]$
\item forward\_15: $[t, \ldots, t+14]$
\item asymm\_past: $[t-11, \ldots, t+3]$
\item asymm\_future: $[t-3, \ldots, t+11]$
\end{itemize}
Each directional window produces a probability distribution with confidence score (maximum probability). We aggregate using confidence-weighted voting (Algorithm~\ref{alg:confweight}).

\begin{algorithm}[!htbp]
\caption{Confidence-Weighted Aggregation}
\label{alg:confweight}
\begin{algorithmic}[1]
\FOR{timestamp $t$ with 7 directional probability distributions}
    \FOR{each direction $d$ ($d = 1$ to $7$)}
        \STATE $\text{confidence}_d = \max(\text{probability\_distribution}_d)$
        \STATE $\text{weighted\_vote}_d = \text{probability\_distribution}_d \times \text{confidence}_d$
    \ENDFOR
    \STATE $\text{final\_probability}_t = \sum(\text{weighted\_vote}_d) / \sum(\text{confidence}_d)$
    \STATE $\text{final\_prediction}_t = \arg\max(\text{final\_probability}_t)$
\ENDFOR
\end{algorithmic}
\end{algorithm}

\begin{figure}[!htbp]
\centering
\includegraphics[width=\columnwidth]{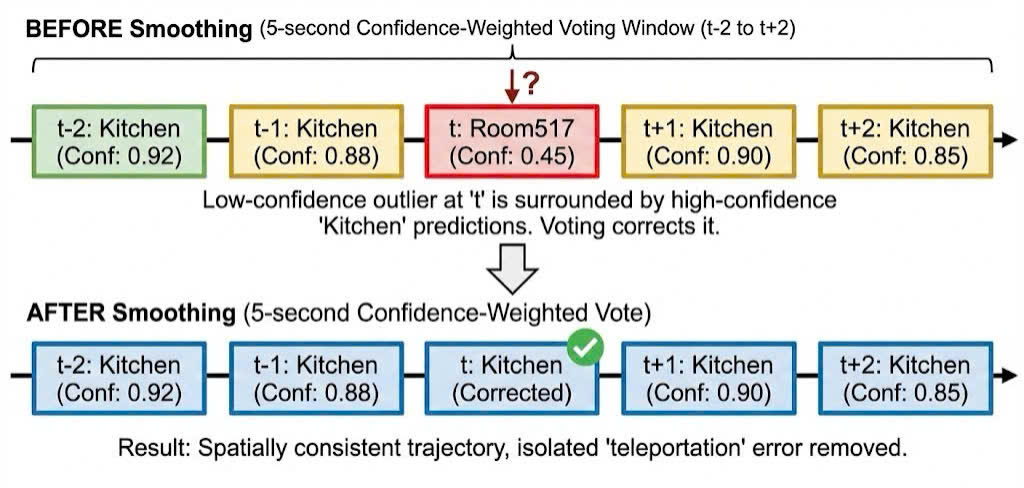}
\caption{5-second temporal smoothing visualization.}
\label{fig:smoothing}
\end{figure}

\paragraph{Phase 4: Temporal Smoothing}
For each prediction at timestamp $t$, we examine a 5-second temporal window $[t-2, t-1, t, t+1, t+2]$ and apply confidence-weighted voting (Algorithm~\ref{alg:smoothing}, Fig.~\ref{fig:smoothing}).

\begin{algorithm}[!htbp]
\caption{Temporal Smoothing}
\label{alg:smoothing}
\begin{algorithmic}[1]
\FOR{each timestamp $j \in [t-2, t-1, t, t+1, t+2]$}
    \STATE $\text{confidence}_j = \max(\text{probability\_distribution}_j)$
    \STATE $\text{weighted\_vote} \mathrel{+}= \text{probability\_distribution}_j \times \text{confidence}_j$
\ENDFOR
\STATE $\text{smoothed\_prediction}_t = \arg\max(\sum \text{weighted\_vote})$
\end{algorithmic}
\end{algorithm}

\FloatBarrier
\subsection{Evaluation Protocol}
We evaluate model performance using macro F1-score, the official metric specified by the ABC 2026 challenge organizers. Macro F1-score computes the F1-score independently for each room class and averages these scores, treating all rooms equally regardless of frequency. This metric is essential for imbalanced indoor localization where balanced performance across all locations is critical for comprehensive care facility monitoring.

We employ 4-fold cross-validation with temporal splitting rather than random partitioning. The dataset spanning four days (April 10-13, 2023) is split such that each fold uses one complete day as the test set and the remaining three days as training. Temporal splitting ensures test data is genuinely unseen and prevents data leakage from highly autocorrelated consecutive BLE readings, simulating realistic deployment where models trained on historical data must generalize to new time periods.

The four folds exhibit substantial variation in data size and class distribution, reflecting natural imbalance in real-world data collection. Table~\ref{tab:folds} presents the detailed characteristics of each fold.

\begin{table}[!htbp]
\caption{Cross-Validation Fold Characteristics}
\label{tab:folds}
\centering
\resizebox{\columnwidth}{!}{%
\renewcommand{\arraystretch}{1.5}
\begin{tabular}{|c|c|c|c|c|c|c|}
\hline
\textbf{Fold} & \textbf{Test} & \textbf{Train} & \textbf{Test} & \textbf{Ratio} & \textbf{Train} & \textbf{Test} \\
 & \textbf{Day} & \textbf{Frames} & \textbf{Frames} & & \textbf{Classes} & \textbf{Classes} \\
\hline
1 & Day 4 & 962,294 & 30,619 & 31.4× & 13 & 13 \\
\hline
2 & Day 3 & 951,141 & 143,401 & 6.6× & 18 & 18 \\
\hline
3 & Day 2 & 747,816 & 333,507 & 2.2× & 15 & 15 \\
\hline
4 & Day 1 & 465,004 & 590,447 & 0.79× & 12 & 12 \\
\hline
\end{tabular}%
}
\end{table}

Fold 1 tests on the smallest dataset (Day 4 with 30,619 frames), providing the largest training set and a highly favorable train/test ratio of 31.4×. Conversely, Fold 4 presents the most challenging scenario: testing on the largest dataset (Day 1 with 590,447 frames) while training on the smallest subset (465,004 frames), resulting in a train/test ratio below 1 (0.79×). This tests the model's ability to generalize when training data is scarce relative to deployment scale. The number of room classes also varies across folds, ranging from 12 to 18 unique locations, reflecting realistic deployment conditions where different days may involve visits to different subsets of rooms.

\FloatBarrier
\section{Results and Analysis}

Table~\ref{tab:results} presents the comprehensive evaluation results comparing traditional machine learning approaches and the proposed DASEL framework across 4-fold cross-validation. All results are reported as macro F1 scores.

\begin{table}[!htbp]
\caption{Macro F1 Scores Across 4-Fold Cross-Validation (F1-F4: Fold 1-4; Var.: Variation)}
\label{tab:results}
\centering
\resizebox{\columnwidth}{!}{%
\renewcommand{\arraystretch}{1.5}
\begin{tabular}{|l|c|c|c|c|c|}
\hline
\textbf{Approach} & \textbf{F1} & \textbf{F2} & \textbf{F3} & \textbf{F4} & \textbf{Mean±Std} \\
\hline
Baseline & 0.2819 & 0.2493 & 0.2665 & 0.3242 & 0.2805±0.0278 \\
\hline
Var. 1 & 0.3009 & 0.2621 & 0.2634 & 0.3327 & 0.2898±0.0293 \\
\hline
Var. 2 & 0.2830 & 0.2486 & 0.2633 & 0.3455 & 0.2851±0.0369 \\
\hline
\textbf{DASEL} & \textbf{0.5114} & \textbf{0.4207} & \textbf{0.4340} & \textbf{0.4082} & \textbf{0.4438±0.0295} \\
\hline
\end{tabular}%
}
\end{table}

The traditional machine learning approaches demonstrate limited and remarkably consistent performance. The baseline method achieved a mean macro F1 score of 0.2805 ± 0.0278. Variation 1 with dominant beacon features achieved the highest traditional ML performance at 0.2898 ± 0.0293, while Variation 2 with signal pattern-based relabeling yielded 0.2851 ± 0.0369. Despite targeting different optimization strategies - feature engineering in Variation 1 versus data augmentation in Variation 2 - all traditional methods remained within a narrow 0.2805 to 0.2898 range. This consistent plateau across fundamentally different techniques suggests these methods have reached a fundamental ceiling imposed by the independence assumption paradigm.

The proposed DASEL framework achieves a mean macro F1 score of 0.4438 ± 0.0295, with individual fold performances ranging from 0.4082 (Fold 4) to 0.5114 (Fold 1). The results demonstrate consistent performance improvements over traditional methods across all folds. The relatively stable standard deviation (0.0295) indicates robust performance across different temporal splits and varying data availability conditions, including the challenging Fold 4 scenario where training data is scarce relative to test data (train/test ratio 0.79×).

DASEL achieves a 53.1\% relative improvement over the best traditional baseline (Variation 1: 0.2898 → 0.4438) and a 58.2\% improvement over the basic traditional approach (Baseline: 0.2805 → 0.4438). The improvement is consistent across all folds: Fold 1 shows a 70.0\% improvement (0.3009 → 0.5114), Fold 2 shows 60.5\% (0.2621 → 0.4207), Fold 3 shows 64.8\% (0.2634 → 0.4340), and Fold 4 shows 22.7\% (0.3327 → 0.4082). Even in the most challenging scenario (Fold 4 with limited training data), DASEL maintains substantial performance gains, demonstrating that the sequential learning paradigm with multi-directional ensemble inference effectively captures temporal movement patterns that traditional independent-window approaches fundamentally cannot model.

\FloatBarrier
\section{Discussion}
The dramatic performance difference between traditional methods and DASEL reveals that the fundamental limitation lies in treating each temporal window as an independent classification problem. Traditional machine learning methods extract statistical features from beacon signals and make predictions independently for each moment. Despite optimization strategies targeting feature engineering (Variation 1) and class imbalance handling (Variation 2), all methods plateaued within 0.2805 to 0.2898 macro F1 score, demonstrating that incremental improvements cannot overcome this independence assumption. Indoor localization is fundamentally a sequential problem - caregivers follow continuous trajectories where location at time $t$ strongly predicts location at time $t+1$, and room transitions produce gradual shifts in beacon patterns. Traditional methods discard this temporal continuity entirely. DASEL's 53.1\% improvement validates that sequential modeling is essential for breakthrough performance. By reconceptualizing indoor localization as a sequential learning problem and treating entire room visits as temporal sequences, DASEL captures movement trajectories, transition dynamics, and temporal patterns that independent-window classification fundamentally cannot model.

The bidirectional GRU architecture with attention mechanism directly addresses the fundamental characteristic of indoor localization data: temporal sequences where human movements follow continuous trajectories rather than independent observations. The core breakthrough comes from the bidirectional GRU layers \cite{b22, b23, b24}, which capture both past context (where the person came from) and future context (where they are going) - this sequential modeling alone accounts for the majority of performance improvement over traditional methods. However, not all timesteps within a sequence are equally informative: during stable room occupancy, beacon patterns provide clear discriminative signals, while doorway passages and transitions produce multi-room ambiguities with overlapping signal characteristics. The attention mechanism \cite{b25, b27} addresses this data characteristic by learning to identify and emphasize key moments within sequences while downweighting noisy transitional periods, effectively simplifying complex sequential patterns into their most informative components. This architectural design - matching the model structure to the inherent temporal dependencies and variable signal quality in real-world movement data - directly translates to the observed 53.1\% performance improvement, demonstrating that the sequential nature of indoor localization requires sequential modeling approaches rather than independent-window classification.

\begin{figure}[!htbp]
\centering
\includegraphics[width=\columnwidth]{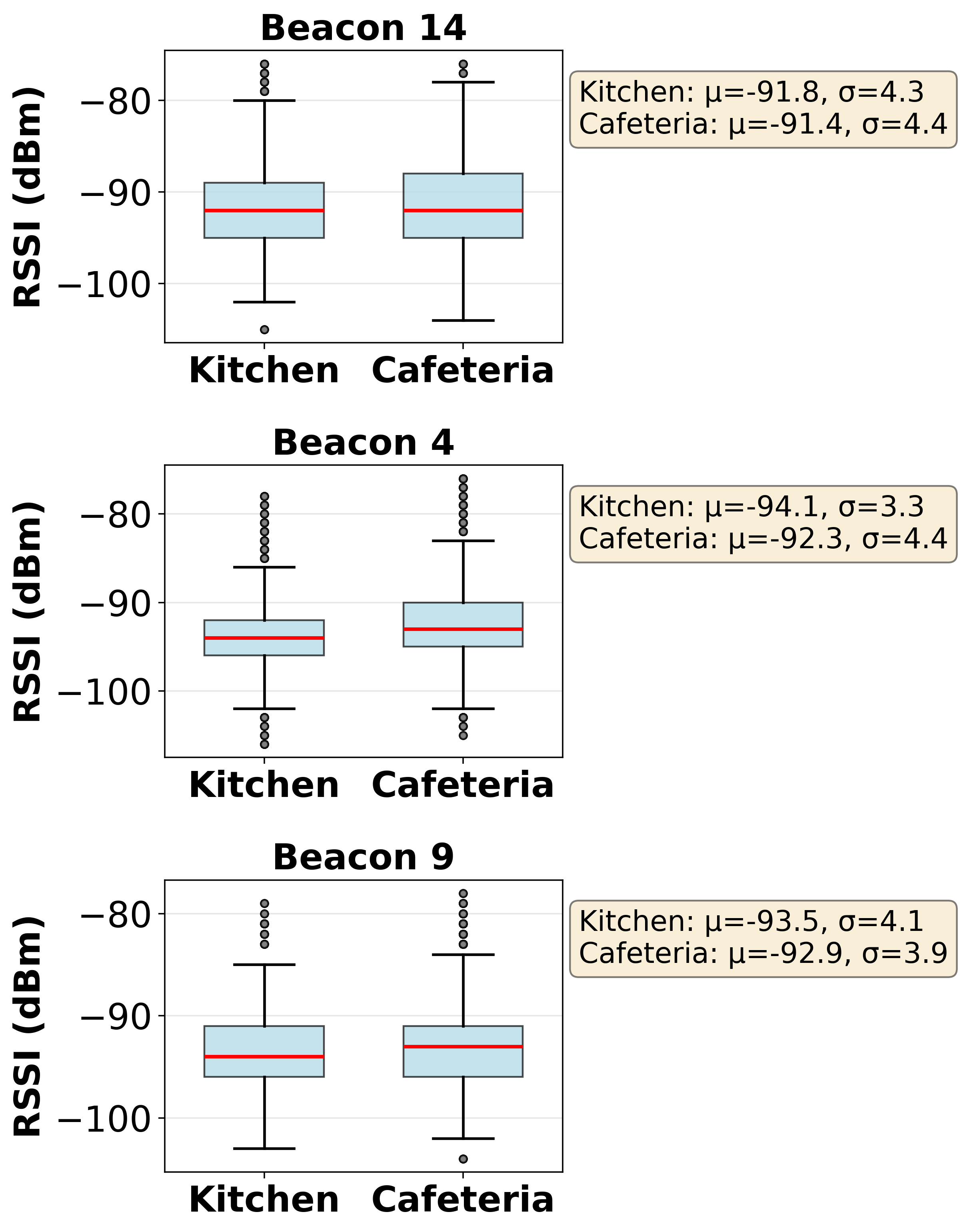}
\caption{RSSI distribution comparison between Kitchen and Cafeteria showing high variance and overlap.}
\label{fig:rssicomp}
\end{figure}

Frequency-based features provide critical practical advantages for real-world deployment. Raw RSSI measurements suffer from severe instability due to multipath signal propagation, human body absorption, interference from other wireless devices, and device-specific hardware characteristics. For example, Fig.~\ref{fig:rssicomp} demonstrates this instability through box plot comparisons of RSSI distributions for three beacons appearing in both Kitchen and Cafeteria, revealing two critical limitations: high intra-room variance (within a single location, signal strength varies substantially), and significant inter-room overlap (mean RSSI values for the same beacon in different rooms differ by less than 2 dBm, smaller than the variance within each room).

\begin{figure}[!htbp]
\centering
\includegraphics[width=\columnwidth]{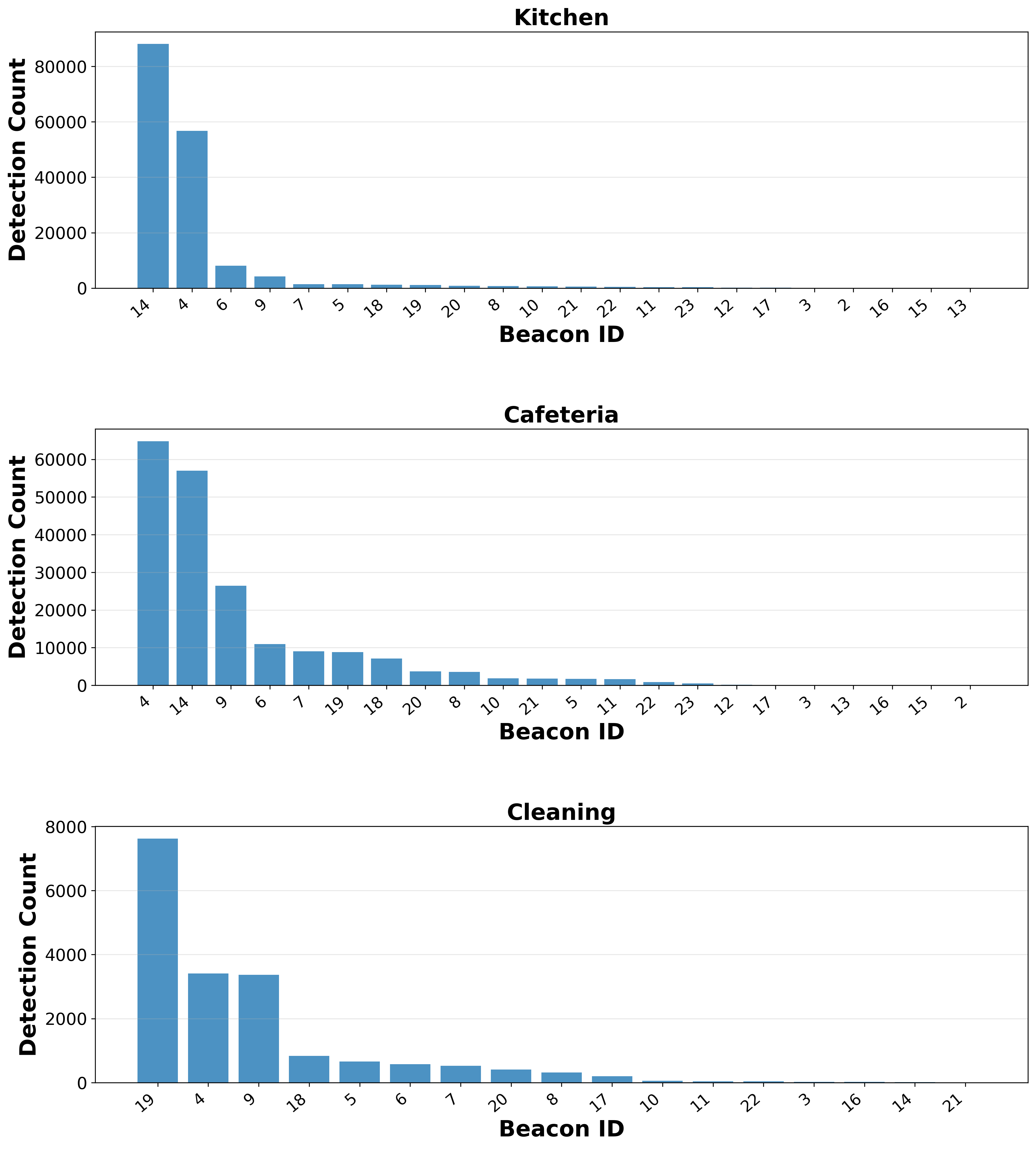}
\caption{Beacon frequency distribution for Kitchen, Cafeteria, and Cleaning showing distinct patterns.}
\label{fig:freqdist}
\end{figure}

In contrast, beacon appearance frequency captures which beacons are detected rather than how strongly, providing a more stable spatial signature. Fig.~\ref{fig:freqdist} illustrates the frequency distribution for three representative rooms, demonstrating that each room exhibits a distinct beacon appearance pattern. Each room shows characteristic detection patterns based on proximity to installed transmitters - nearby beacons appear frequently while distant beacons appear rarely or not at all. These presence patterns remain consistent despite environmental factors that destabilize individual RSSI values.

The frequency representation offers natural normalization (values between 0 and 1) without device-specific calibration, eliminating a major practical barrier to deployment across heterogeneous mobile devices.

The two-level hierarchical ensemble addresses distinct challenges that enhance prediction quality beyond the core sequential modeling framework. Multi-seed training with five different random initializations reduces variance by averaging predictions across models that learned slightly different patterns from unique initialization trajectories. This ensemble strategy provides robustness against unlucky initializations that might converge to poor local minima, ensuring evaluation metrics reflect the architecture's true capability rather than initialization luck. The multi-directional sliding window strategy addresses a fundamental inference challenge: during training, the model learns from sequences with known room boundaries, but during deployment these boundaries are unknown. For any timestamp requiring prediction, we don't know where it falls within its actual room visit - at the beginning, middle, or end. No single window configuration optimally captures all positions. Backward-looking windows fail when predicting near sequence start, forward-looking windows fail near sequence end, and centered windows struggle at boundaries. By employing seven complementary temporal perspectives, at least some windows will be well-aligned with clean signals regardless of true positional context.

\begin{figure}[!htbp]
\centering
\includegraphics[width=\columnwidth]{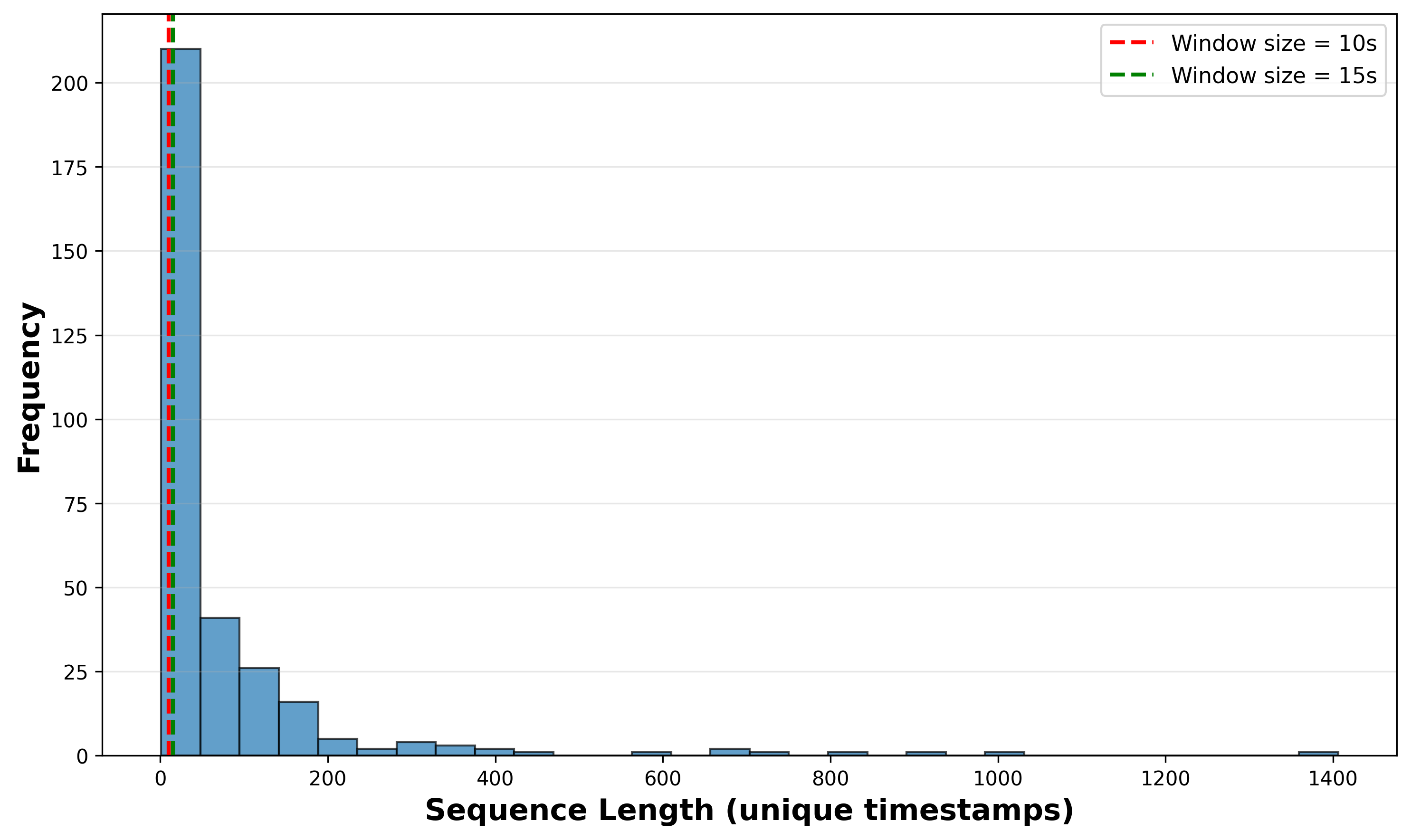}
\caption{Training sequence length distribution showing concentration between 10-200 seconds.}
\label{fig:seqdist}
\end{figure}

The choice of window sizes (10 and 15 seconds) is empirically grounded in the training data's sequence length distribution. Fig.~\ref{fig:seqdist} shows the distribution of training sequence lengths, revealing that the majority of room visits are concentrated between 10 and 200 seconds. The 10-second window captures the lower end of typical visit durations, ensuring even brief passages receive adequate temporal context without excessive contamination. The 15-second extended windows offer additional context for longer visits while remaining well below the average sequence length, minimizing risk of spanning multiple room transitions.

Confidence-weighted aggregation allows well-aligned windows with high-confidence predictions to dominate while poorly-aligned windows contaminated by transitions contribute less, creating a self-regulating ensemble.

The 5-second temporal smoothing serves as final post-processing that enforces spatial and temporal consistency. Even after multi-directional ensemble aggregation, isolated prediction errors can occur where a single timestamp is incorrectly classified as a distant room despite surrounding predictions consistently indicating a nearby location. These isolated errors are often physically implausible - people don't teleport instantaneously across buildings. The smoothing window examines predictions within a local temporal neighborhood and applies confidence-weighted majority voting, effectively eliminating teleportation errors while preserving legitimate room transitions that show sustained directional change. The 5-second window balances catching isolated errors (people typically remain in rooms longer than 5 seconds during meaningful visits) while preserving genuine transitions between adjacent rooms (which can occur within 5-10 seconds).

DASEL's balanced macro F1 score indicates reliable performance across all room classes including rarely-visited locations, which is essential for comprehensive care facility monitoring where accurate localization in individual patient rooms is as critical as tracking in common areas. The frequency-based representation combined with temporal smoothing produces spatially consistent predictions matching realistic human movement patterns. These characteristics - balanced performance, device robustness, and spatial consistency - make DASEL deployable in real-world settings without extensive system tuning. However, the two-level ensemble requires multiple forward passes per timestamp, which may challenge real-time deployment on resource-constrained mobile devices. Future work could explore model compression techniques such as knowledge distillation to maintain accuracy while reducing computational demands, adaptive window sizing based on detected movement patterns, and incorporating facility layout information to reduce physically implausible errors and improve transition modeling between adjacent rooms.

\FloatBarrier
\section{Conclusion}
This research demonstrates that traditional machine learning approaches to BLE-based indoor localization face fundamental performance limitations rooted in their independence assumption rather than implementation details. Our proposed DASEL framework achieves breakthrough performance by reconceptualizing indoor localization as a sequential learning problem, explicitly modeling entire room visits as temporal sequences through bidirectional recurrent networks with attention mechanisms. The integration of frequency-based features for RSSI robustness, two-level hierarchical ensemble for variance reduction and positional uncertainty handling, and confidence-weighted temporal smoothing for spatial consistency creates a comprehensive system that captures movement trajectories, entry/exit patterns, and dwell-time dynamics that traditional methods discard. The resulting 53.1\% improvement demonstrates that the sequential nature of indoor localization is not a secondary consideration but the fundamental property that must be modeled to achieve substantial performance gains. DASEL's balanced performance across all room classes and robustness to device heterogeneity make it practically deployable in real-world care facilities, establishing that breakthrough results in BLE-based indoor localization require treating human movement as the continuous temporal process it fundamentally is.

\section*{Acknowledgment}
The authors would like to thank Kyushu Institute of Technology for providing the dataset in conjunction with ABC 2026.

\balance

\end{document}